%% file: main.tex
\documentclass[runningheads]{llncs}
\usepackage[misc]{ifsym}
\usepackage{bm}
\usepackage{bbm}
\usepackage{graphicx}
\usepackage{algorithm}
\usepackage{algpseudocode}
\usepackage{booktabs}
\usepackage{bbding}
\usepackage{amssymb}
\usepackage{multirow}
\usepackage[pagebackref,breaklinks,colorlinks]{hyperref}
\usepackage[capitalize]{cleveref}
\renewcommand\UrlFont{\color{blue}\rmfamily}

\begin{document}
\title{UDD: Dataset Distillation via Mining Underutilized Regions}
%
%
\author{Shiguang Wang\inst{1}\and
Zhongyu Zhang\inst{2} \and
Jian Cheng\inst{1,}\textsuperscript{\Letter}}
\authorrunning{F. Author et al.}
%
\institute{University of Electronic Science and Technology of China, No.2006, Xiyuan Ave, West Hi-Tech Zone, Chengdu, China \\
\email{xiaohu\_wyyx@163.com, chengjian@uestc.edu.cn}\\
\and
Youtu Lab, Tencent Technology Co.Ltd, Shanghai, China \\
\email{zzy28047@163.com}}
\maketitle              
\begin{abstract}
Dataset distillation synthesizes a small dataset such that a model trained on this set approximates the performance of the original dataset. Recent studies on dataset distillation focused primarily on the design of the optimization process, with methods such as gradient matching, feature alignment, and training trajectory matching. However, little attention has been given to the issue of underutilized regions in synthetic images.
In this paper, we propose UDD, a novel approach to identify and exploit the underutilized regions to make them informative and discriminate, and thus improve the utilization of the synthetic dataset. Technically, UDD involves two underutilized regions searching policies for different conditions, i.e., response-based policy and data jittering-based policy. Compared with previous works, such two policies are utilization-sensitive, equipping with the ability to dynamically adjust the underutilized regions during the training process. Additionally, we analyze the current
model optimization problem and design a category-wise feature contrastive loss, which can enhance the distinguishability of different categories and alleviate the shortcomings of the existing multi-formation methods. 
Experimentally, our method improves the utilization of the synthetic dataset and outperforms the state-of-the-art methods on various datasets, such as MNIST, FashionMNIST, SVHN, CIFAR-10, and CIFAR-100. For example, the improvements on CIFAR-10 and CIFAR-100 are 4.0\% and 3.7\% over the next best method with IPC=1, by mining the underutilized regions.

\end{abstract}
\section{Introduction}
 \label{sec:intro}
 A very simple way to guarantee the outstanding performance of 
deep neural network (DNN) is to use a large number of training data (ignore the bias of the dataset temporarily). Unfortunately, training on the large-scale dataset will cost enormous computational resources and a good while.
\textbf{Dataset distillation} (or called \textbf{dataset condensation }), proposed by Wang et al. \cite{DD} in the seminal 2018 paper, aims to alleviate the cumbersome training process by constructing a small training set.
Related but orthogonal to knowledge distillation \cite{KD} where teacher and student models are expected to have similar responses on the same data, dataset distillation keeps the model consistent but encapsulates the knowledge of the entire training dataset. Typically, it compresses the original real dataset containing thousands to millions of images into a small number of the synthetic dataset. 
Dataset distillation has been successfully applied in various applications \cite{Flexible,Kernel,Infinitely,Generative-Teaching-Networks,FLSD,X-Ray,SecDD,IDC}, such as continual learning, neural architecture search, and privacy-preserving ML.

Unlike classic data compression methods which only select valuable data for model training, the core idea of dataset distillation is not to filter the subset but to synthesize, which means the synthesis process is more critical. 
Recently, dataset distillation methods \cite{DD,DC,CAFE,MTT} mainly focus on the study of optimization scheme in the synthesis process.
The seminal approach \cite{DD} proposes a meta-learning-based method that utilizes a gradient-based hyperparameter optimization \cite{GHO} scheme for data and model nested loop update.
To alleviate the high time consumption caused by nested loop optimization, Zhao et al. \cite{DC} regards dataset condensation as a gradient matching problem between the gradients of training networks that are trained on the original and the synthetic data respectively. 
As for optimization formulation in the synthesis process, feature alignment \cite{CAFE} and training trajectory \cite{MTT} are proven to be effective condensation information.

\begin{figure}[t]
	\centering
\includegraphics[width=0.75\linewidth]{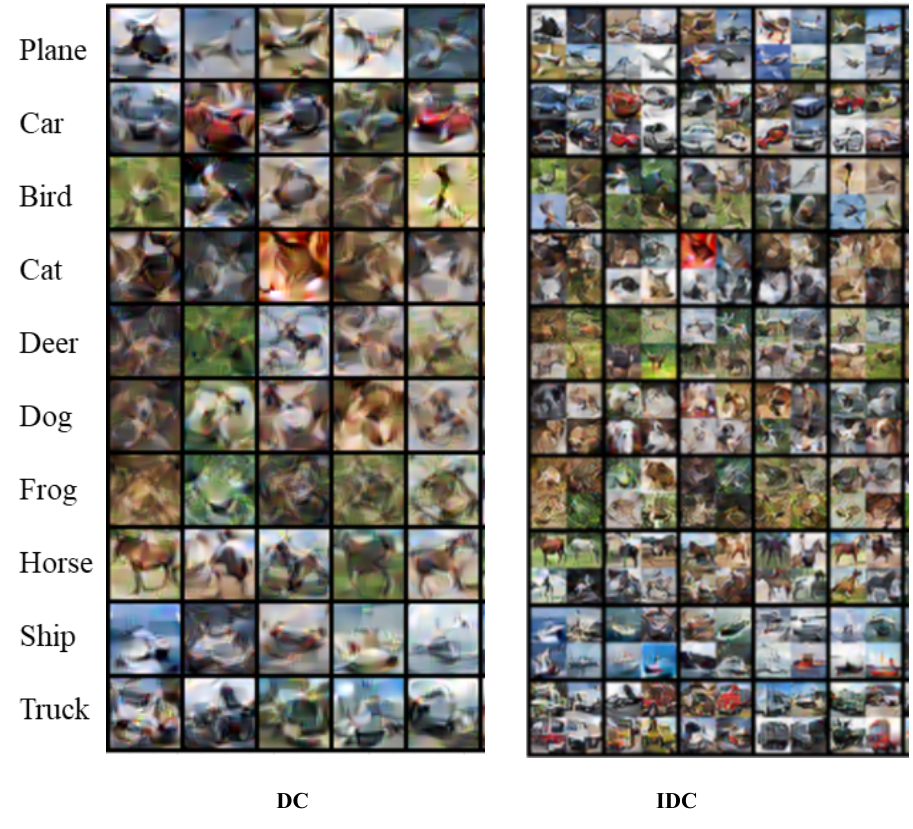}
		\caption{Examples of the synthetic images from 32 $\times$ 32 CIFAR-10 dataset.
			 \textbf{Left}: synthetic images generated by DC \cite{DC}, which has many underutilized regions (concentrated around the border).
             \textbf{Right}: synthetic images generated by IDC \cite{IDC}, that the unbalanced information distribution is partially alleviated.}
		\label{fig:utilization_vis}
   \vspace{-0.7em}
\end{figure}

Despite their great success in various applications, most of the existing dataset distillation methods neglect the problem of the unbalanced information distribution in synthetic images, resulting in many underutilized regions, which can be shown in \cref{fig:utilization_vis}.
However, when a smaller image per class(IPC) comes to us, e.g., IPC=1, it becomes crucial to utilize each area of the synthetic image fully.
While IDC \cite{IDC} has attempted to solve this problem by dividing each synthetic image into a fixed number of sub-images for training models, it treats all sub-images equally, ignoring the most underutilized regions and decreasing the utilization of the condensed data elements. Additionally, the fixed and independent partitioning of each synthetic image means that sub-images have limited opportunities to share information. Consequently, we are prompted to ask: how can we identify and improve the underutilized regions in synthetic images?

In this work, we propose a novel Utilization-sensitive Dataset Distillation framework, termed \textbf{UDD}, to mine the underutilized regions of the synthetic data. 
Specifically, we first generate multiple candidate regions with various size for initialization. These candidate regions are partially overlapped and are allowed to share information.
Based on these candidate regions, we design two underutilized regions searching policy, to exploit the underutilized regions, namely \textbf{response-based policy} and \textbf{data jittering-based policy}. Response-based policy takes activation and gradient into account, built upon the insight that uninformative regions have smaller gradient and activation values.
Data jittering-based policy concerns the changing magnitude of the activation values after data jittering.
The motivation behind it is that the network is less sensitive to jitter in uninformative regions such as background areas.
The suggested two policies are designed for different conditions, i.e., time-saving first and accuracy first, to assess how well each part of the image is utilized. 
Subsequently, these underutilized regions are used to form the synthetic training images with the original synthetic image for model optimization and utilization improvement.

Moreover, we have observed that the synthetic dataset trained on
a network that adopts the real dataset as the training data exhibits poor distinguishability of features between different categories.
This lack of discrimination hinders the network's ability to accurately classify objects, thus limiting its usefulness.
To tackle this issue, we design a Category-wise Feature Contrastive (CFC) loss that encourages larger discrimination between the different categories in the synthetic dataset. The CFC loss is combined with the gradient matching loss to form the synthetic image optimization loss.
Our approach enables networks to take advantage of real datasets during training while still producing synthetic datasets that are highly discriminative and effective for the downstream task.

Experiments are carried out with standard datasets, including MNIST, SVHN, CIFAR-10, and CIFAR-100. The results yielded by our method outperform existing dataset distillation methods as well as coreset selection methods. Compared with state-of-the-art methods, we typically achieved 54.6\% on CIFAR-10 and 27.9\% on CIFAR-100 with a single image per class, obtaining 4.0\% and 3.7\% performance improvements, respectively. We then propose an evaluation metric $mUE$ to evaluate the utilization of the synthetic dataset, which delivers a quantization observation of distribution and utilization.

The main contribution of this work can be summarized as (1) 
We introduce a novel dataset distillation framework, called the Utilization-sensitive Dataset Distillation (UDD) framework, which can handle demanding scenes, such as smaller IPC, by dynamically adjusting the utilization of underutilized regions.
(2) We construct two underutilized regions searching policies for different conditions, which can be used to efficiently identify underutilized regions in the synthetic dataset and adjust them during the training process.
(3) We address the current limitation of the model optimization problem by introducing a Category-wise Feature Contrastive (CFC) loss, which strengthens the distinguishability of features across different categories of the synthetic dataset.
(4) Comprehensive experiments are conducted to demonstrate the effectiveness of our proposed UDD. Furthermore, we give a quantization analysis of the distribution and utilization of the synthetic dataset.

\section{Utilization-sensitive Framework}
\begin{figure*}[t]
	\centering
	\includegraphics[width=0.95\linewidth]{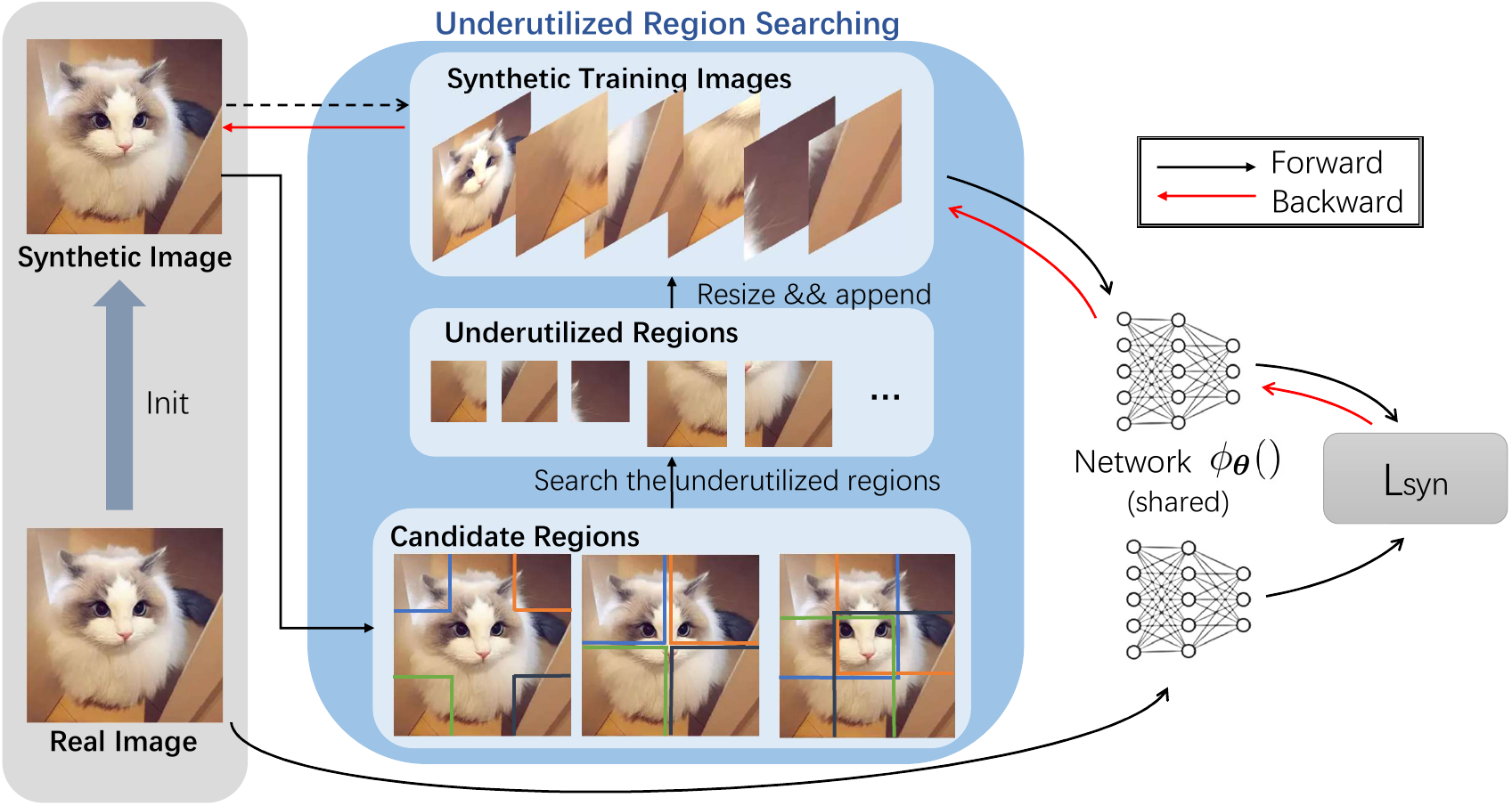}
	\caption{Illustration of the proposed dataset distillation framework. 
		\textbf{Underutilized Region Searching}: the module for mining the underutilized regions, which will be optimized with the original synthetic data in an end-to-end fashion.
		\textbf{Candidate Regions}: illustration of dividing each synthetic image into multiple regions. 
		To facilitate observation, the regions in each sub-image are slightly staggered.
		\textbf{Underutilized Regions}: the underutilized regions will be resized to the original size by bilinear upsampling and combined with the original synthetic image to compose the synthetic training images.}
	\label{fig:framework}
	\vspace{-0.5em}
\end{figure*}
\subsection{Overview}
The goal of dataset distillation is to synthesize a small, synthetic dataset $\mathcal{S} = \{(x^{syn},y^{syn})\}$ that the model trained with this dataset approximates the results with the large, real dataset $\mathcal{T} = \{(x^{real},y^{real})\}$. 
Most of previous approaches directly optimize each data element, without imposing regularity conditions on the the synthesize images. 
However, it is not tailored for improving the utilization of limited area of the synthetic images. 
In this work, we propose the utilization-sensitive dataset distillation framework to mine these underutilized regions, which is shown in \cref{fig:framework}.
Before the optimization of data distillation, we randomly sample a batch from the real dataset to initialize the synthetic dataset, which could better discriminate the utilization of each region comparing to initialized with noise.
At each distillation step, the synthetic image is divided into multiple sub-images, which is known as \textbf{candidate regions}.
Then, an underutilized region searching approach is proposed to capture the utilization of each region of the image, which is illustrated in \cref{sec:by_response} and \cref{sec:by_jitter} in detail. 
The underutilized regions are resized to the original size of the synthetic image to form the \textbf{synthetic training images} $\mathcal{S}^{train}$ with the original synthetic image.
Finally, the $\mathcal{S}^{train}$ are updated by $L_{syn}$, which is composed of 1). an MSE loss for minimizing the two gradient sets of the networks that are computed over a largely fixed training set and a learnable condensed set respectively, and 2). a contrastive loss for maximizing the agreement between images with the same category.
The \textbf{proxy network} $\phi_{\theta}()$ was updated by minimizing the cross-entropy loss on the real dataset. Note that 1). the underutilized regions is differentiable, and these regions in the original synthetic data can be improved in an end-to-end fashion, and 2). the underutilized regions is dynamic adjusted along with the training process, which avoids being dominated by some fixed regions, and 3). the training process of proxy network $\phi(\theta)$ is not illustrated in \cref{fig:framework} for simplicity and the full optimization process can be accessed in \cref{alg:pipeline}. 

\subsection{Candidate Regions Generation}
In this section, we describe how to generate candidate regions for initialization.
Specifically, given a synthetic image with size of $\mathit{H} \times \mathit{W}$,
we divide it into multiple regions $\mathit{R} = [\mathit{R}_1, \mathit{R}_2,...,\mathit{R}_N] $ using a series of predefined dimensions. 
As illustrated in Candidate Regions in \cref{fig:framework}, 
these candidate regions are divided into three groups with dimensions of $(\mathit{H}/3\times\mathit{W}/3),(\mathit{H}/2\times\mathit{W}/2)$ and $(2\mathit{H}/3\times2\mathit{W}/3)$ respectively, corresponding to small, medium and large regions.
Let's take group $(\mathit{H}/3\times\mathit{W}/3)$ as an example, 4 rectangular windows with dimensions of $(\mathit{H}/3\times\mathit{W}/3)$ are located on the top-left, top-right, bottom-left, and bottom-right corner of the synthetic image, respectively. 
The location of windows is based on the insight that the corner of the synthetic images is hard to optimize due to the features of convolutional neural network. Similar conclusion can also be visualized in \cref{fig:utilization_vis}. 
Then, we operate these window groups on the synthetic image to obtain diverse candidate regions with various sizes and information sharing.
Notably, the sliding window is not adopted for saving computing resources. 

Next, we present our approach for searching underutilized regions from candidates based on different observations, which is depicted in \cref{sec:by_response} and \cref{sec:by_jitter}, respectively.

\subsection{Underutilized Regions Searching by Response-based Policy}
\label{sec:by_response}
\begin{figure}[t]
	\centering
\includegraphics[width=0.85\linewidth]{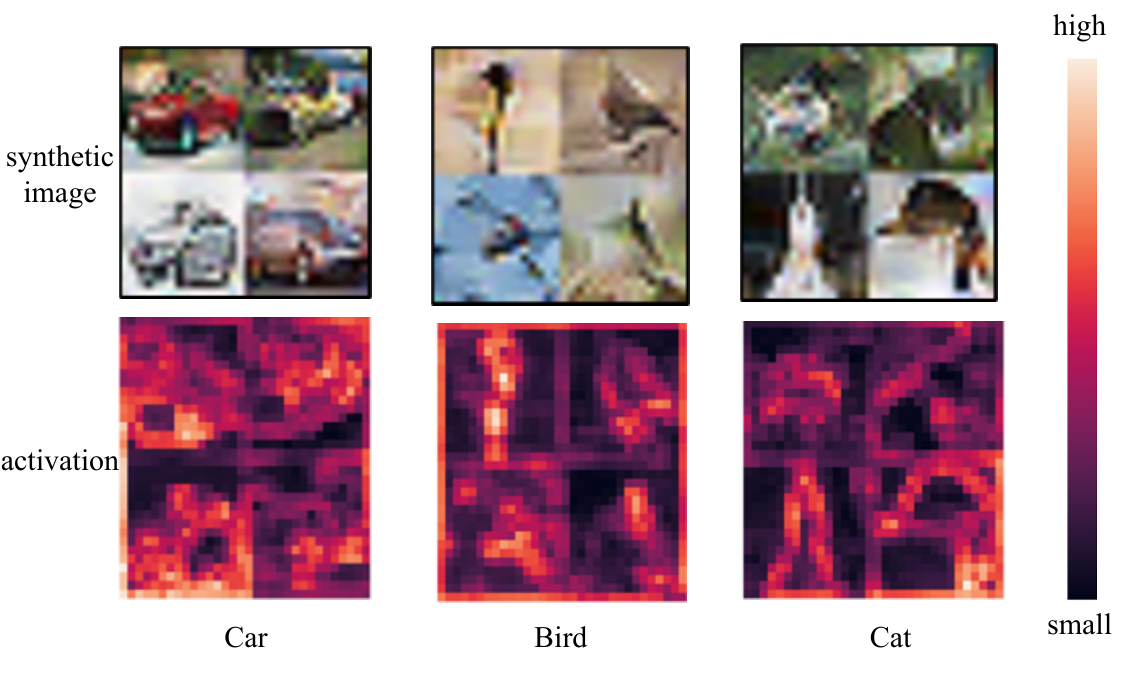}
		\caption{Examples of the synthetic images with their activation maps from 32 $\times$ 32 CIFAR-10 dataset. The synthetic images are generated by IDC \cite{IDC}.}
		\label{fig:fig-act}
   \vspace{-0.5em}
\end{figure}

Activation and gradient are two essential components of network optimization, where their response reflects the intensity of neural activation and updating during training, respectively. 
Inspired by salient object detection \cite{borji2019salient,borji2015salient}, we assume that the higher the response of activation and (or) gradient, the higher the utilization of the candidate regions. 
As a preliminary step, we conduct experiments to observe the activation response of different areas of the synthetic image.
As depicted in \cref{fig:fig-act}, the full-utilized regions have a high response while the underutilized regions have a low response.

Given the candidate regions $\mathit{R} = [\mathit{R}_1, \mathit{R}_2,...,\mathit{R}_n,...,\mathit{R}_N]$ and the network $\phi_{\theta}()$ with $L$ layers, we are supposed to get the corresponding response $\Lambda_n^l$ of the $l$-th layer, resize and append the lowest $P$ regions $\mathit{R}^*$ to the synthetic training images $\mathcal{S}^{train}$. 
Here, we neglect $l$ for simplicity.
The indicator $\Lambda_n$ can be expressed as:
\begin{equation}
	\Lambda_n = F(\mathit{R}_n; \phi_{\theta}()).
\end{equation}

\begin{figure}[t]
	\centering
\includegraphics[width=0.85\linewidth]{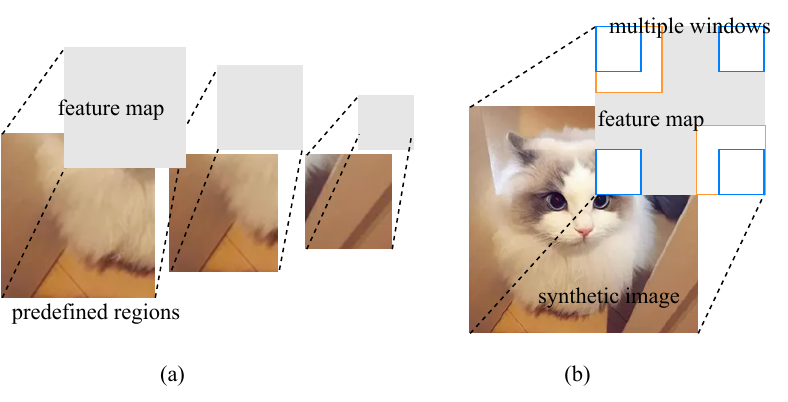}
		\caption{Different schemes for addressing predefined regions (candidate regions) and their corresponding response. (a). Different predefined regions are built, and the network is run at all regions. (b). Different windows are run on the feature map.}
		\label{fig:fig-response-imp}
   \vspace{-0.5em}
\end{figure}

A naive approach to get $\Lambda_n$ is serially running $N$ candidate regions on the network, which is shown in \cref{fig:fig-response-imp} (a). 
However, it's time-consuming. 
Instead, our scheme \cref{fig:fig-response-imp} (b) runs the whole synthetic image one time and gets its response. 
Then, multiple windows (maybe downsample accordingly) corresponding to each candidate region are placed on it, which avoids enumerating multiple candidate regions.

\noindent\textbf{Activation Ranking.}
In this setting, activation is regarded as the response metric. 
We first forward the network $\phi_{\theta}()$ and get all the activations of candidate regions. Then, the activations are averaged to get the final $\Lambda_n$.

\noindent\textbf{Gradient Ranking.}
Regions with large gradients produce dominant gradients over the network parameters. 
Those representative but easy regions may be overlooked during the synthesis process, which will aggravate their underutilization.
Thus, we design a novel but simple gradient ranking approach that focuses on improving the utilization of low gradient ranking regions.
Similarly, We forward and backward the network $\phi_{\theta}(\cdot)$ and get all the gradients of candidate regions.
Then, we rank the average of gradients for each region, resize and append the lowest $P$ regions into $\mathcal{S}^{train}$.

\subsection{Underutilized Regions Searching by Data Jittering-based Policy}
\label{sec:by_jitter}
In the process of network optimization, data jittering is an efficient augmentation technique to improve the robustness of the input data. 
Here, We propose a novel utilization definition based on data jittering that takes advantage of the network's insensitivity to small perturbations in uninformative regions, such as background areas. Specifically, we define utilization as the difference between the mean activation value of the jittered image and the original image. This approach allows us to effectively identify and improve the underutilized regions in synthetic images.

Specifically, given a candidate region $\mathit{R}_n$, we sample a set of jittered regions
$\hat{\mathit{R}}_n = [\hat{\mathit{R}}_{n,1}, \hat{\mathit{R}}_{n,2}, ...,\hat{\mathit{R}}_{n,m}, ..., \hat{\mathit{R}}_{n,M}]$ by using different seeds, which is formulated as follows:
\begin{equation}
\hat{\mathit{R}}_{n,m} = jitter(\mathit{R}_n).
\label{eq-jitter}
\end{equation}

Typical $jitter$ function including: 
gaussian noise, salt and pepper noise, and uniform noise \cite{Gaussian}.
We choose gaussian noise as the default $jitter$ function, which is a noise with a probability density function of a normal distribution (also called Gaussian distribution).
After getting the jittered regions, we follow the scheme in \cref{sec:by_response} to obtain their response and denote $\hat{\Lambda}_{n,m}$ as the activation of jittered region $\hat{\mathit{R}}_{n,m}$.

Then, we define the utilization $\sigma_n$ of the candidate region $\mathit{R}_n$ as the difference between the mean value of the jittered activations $\bar{\Lambda}_n$ and the original activation $\Lambda_n$:
\begin{equation}
\bar{\Lambda}_n = \frac{1}{M}\sum_{m=1}^M\hat{\Lambda}_{n,m},
\label{eq-mean}
\end{equation}
\begin{equation}
\sigma_n = |\Lambda_n - \bar{\Lambda}_n|.
\label{eq-sigma}
\end{equation}

The lowest $P$ regions with $\sigma$ are regarded as underutilized regions. 
These underutilized regions will be resized to the size of the synthetic image by bilinear upsampling and combined with the original synthetic image
to compose the synthetic training images $\mathcal{S}^{train}$. 

We propose two different policies for searching underutilized regions: the response-based policy and the data jittering-based policy, each based on different observations. The response-based policy is more time-efficient as it only requires one extra network execution. However, its robustness may be lower at the beginning of training since the gradient and activation values are highly dependent on the model. On the other hand, the data jittering-based policy is more robust since it considers the sensitivity of the network to jitter in uninformative regions. However, it may require more computational resources due to the need for data augmentation. Overall, each policy is designed for different conditions and their effectiveness.

\subsection{Improved Model Optimization}
\begin{figure}[t]
	\centering
	\includegraphics[width=0.55\linewidth]{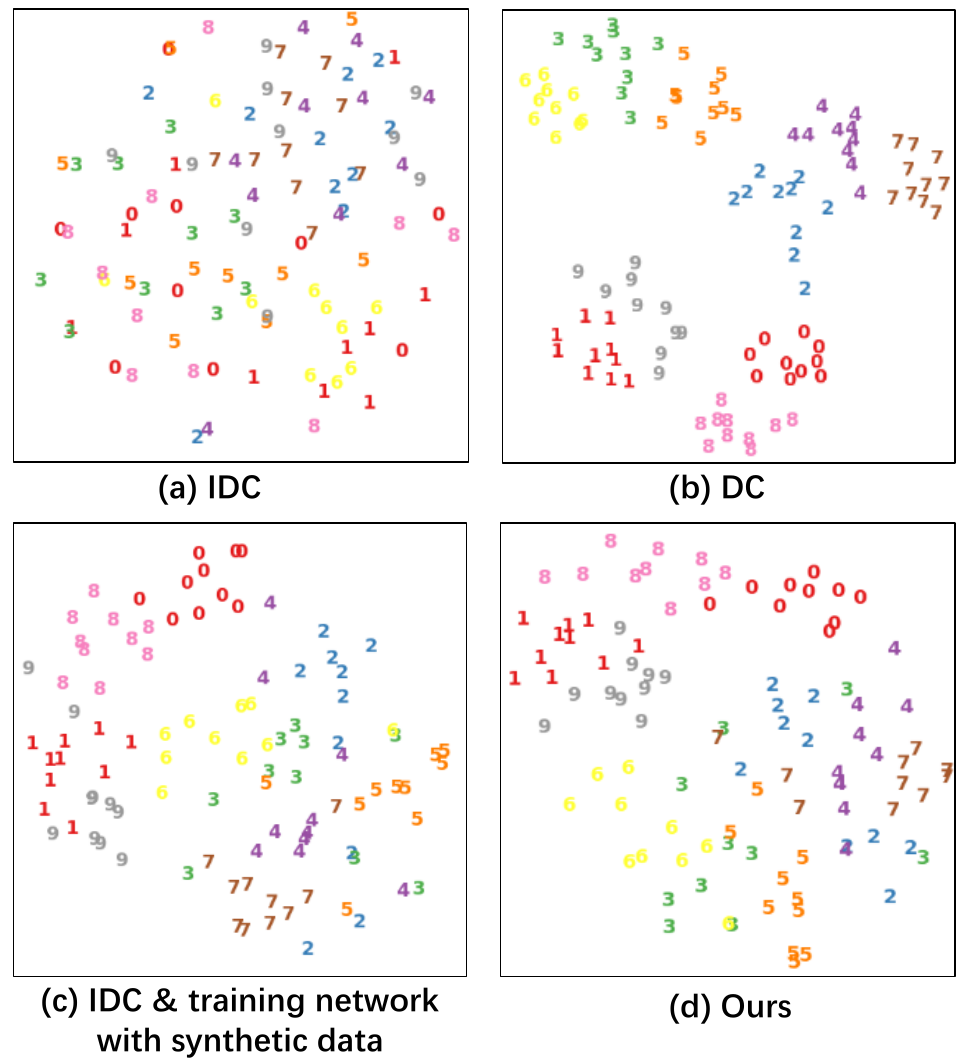}
	\caption{The data distribution of synthetic images from CIFAR-10, which trained on a randomly initialized neural network from scratch. The ``numbers'' with different colors represent the category corresponding to the number in CIFAR-10. For example, ``0'', ``1'', and ``2'' respectively represent ``airplane'', ``automobile'' and ``bird''.}
	\label{fig-tsne}
	\vspace{-0.5cm}
\end{figure}

\noindent\textbf{Problem with Model Optimization.}
Most existing dataset distillation methods use \cref{eq-net_train} to train network weights $\theta$ on the condensed dataset $\mathcal{S}$, where $\eta$ denotes the learning rate and $\theta_t$ denotes the network weights at $t$-th training step. $l(\theta; \mathcal{S})$ denotes the training loss for weight $\theta$ trained on synthetic dataset $\mathcal{S}$. As mentioned in IDC \cite{IDC}, the above optimization approach has two shortcomings, namely chicken-egg problem and gradients vanishment. IDC proposed to train the network using the real dataset instead of the synthetic dataset as \cref{eq-net_train_t} to overcome these drawbacks, resulting in a better performance.
\begin{equation}
\label{eq-net_train}
	\theta_{t+1} = \theta_t - \eta\nabla_\theta l(\theta_t; \mathcal{S}),
\end{equation}
\begin{equation}
	\label{eq-net_train_t}
	\theta_{t+1} = \theta_t - \eta\nabla_\theta l(\theta_t; \mathcal{T}).
\end{equation}

However, we investigate that the synthetic dataset trained on a network that adopts the real dataset as the training data has poor distinguishability of features between different categories. To analyze this problem more intuitively, we utilize t-SNE \cite{tsne} to visualize the features of synthetic images condensed by IDC, DC \cite{DC}. As illustrated in \cref{fig-tsne}, the features of various categories of IDC are mixed, while images condensed by DC can be well differentiated between categories. Both of the above methods use gradient matching, but DC trains the network with synthetic images. So we assume that this phenomenon is brought about by training the model with real data. To verify the assumption, we condense the dataset by IDC but train the network using synthetic images. As shown in \cref{fig-tsne} (c), the method that is consistent with IDC except for training the network using synthetic images has a greatly improved on the discrimination of each category compared with IDC.
However, the performance of this method decreases significantly compared with training the network using real images. There raises a question: how to avoid the above two shortcomings while strengthening the distinguishability of features to further boost the performance?

\noindent\textbf{Solution.}
To tackle this issue while taking advantage of training the network using the real dataset, we design a Category-wise Feature Contrastive (CFC) loss, which is inspired from SimCLR \cite{SimCLR}. Specifically, we sample a batch of synthetic training data $\mathcal{S}_c$ with the same category label c. Then, we embed the batch of synthetic training data using network $\phi_{\theta}()$ and obtain their features 
$F_c = [f_{c,1}, f_{c,2}, ..., f_{c,|\mathcal{S}_c|}] = \phi_{\theta}(\mathcal{S}_c)$. The CFC loss function is defined as
\begin{equation}
	\label{eq-CFC}
	\mathcal{L}_{c} = -log \frac{exp(sim(\bar{f}_c, \bar{F}_c)/\tau)}{\sum_{i=1}^{C} \mathbbm{I}_{[i\neq c]}exp(sim(\bar{f}_c, \bar{F}_i)/\tau)},
\end{equation}
where $\mathbbm{I}_{[i\neq c]} \in \{0,1\} $ is an indicator function evaluating to 1 if $i\neq c$, $\tau$ denotes a temperature parameter and 
$sim(\bm{u},\bm{v}) = \bm{u}^T\bm{v}/\left \|\bm{u}\right \|\left \|\bm{v}\right \|$. 
$\bar{f}_c$ is the mean feature of the current batch and 
$\bar{F}_c$ is the mean feature of category c using the moving average, which is formulated as
\begin{equation}
	\label{eq-moving average}
	\bar{F}_c^{t+1} = \alpha\bar{F}_c^{t} + (1-\alpha)\bar{f}_c^t,
\end{equation}
where $\bar{f}_c^t$ denotes the mean feature of category c at $t$-th training step, $\alpha$ denotes a moving average parameter that controls the moving range.
As illustrated in \cref{fig-tsne}, our method obtains a discrimination effect similar to the method that is consistent with IDC except for training the network using synthetic images.
\begin{algorithm}[t] 
	\caption{UDD: Dataset Distillation via
Exhausting Underutilized Region} 
	\label{alg:pipeline} 
	\small
	\textbf{Input}: $\mathcal{T}$: real data; $\mathcal{S}$: synthetic data. \\
	\textbf{Input}: $C$: number of categories in the current dataset. \\
	\textbf{Input}: $l()$: loss function for model parameter update.
	
	\begin{algorithmic}[1]
		\While{not converged}
		\State Randomly initialize $\theta$.
		\For{t = 1 $\rightarrow$ training steps}
		\For{c = 1 $\rightarrow$ C}
		\State Sample a batch of $\mathcal{T}_c \sim \mathcal{T}$ and $\mathcal{S}_c \sim \mathcal{S}$.
		\State Generate the synthetic training images $\mathcal{S}^{train}_c$ by \cref{sec:by_response} or \cref{sec:by_jitter}.
		\State Update $\mathcal{S}^{train}_c$ by \cref{eq-loss}.
		\EndFor
		\State Sample a mini-batch $T \sim \mathcal{T}$.
		\State Update $\theta_{t+1} = \theta_t - \eta\nabla_{\theta} l(\theta_t; T)$.
		\EndFor
		\EndWhile
	\end{algorithmic} 
	\textbf{Output}: synthetic data $\mathcal{S}$.
\end{algorithm}

Subsequently, we embed real data and synthetic training data using network $\phi_{\theta}()$ and obtain their training loss $\mathcal{L}^\mathcal{T}$ and $\mathcal{L}^\mathcal{S}$. Then, MSE is applied to calculate the gradient matching loss $\mathcal{L}_g$ as:
\begin{equation}
	\mathcal{L}_g = \sum_{\theta_i \in \theta} |\nabla_\theta\mathcal{L}^\mathcal{T}(\theta_i) - \nabla_\theta\mathcal{L}^\mathcal{S}(\theta_i) |^2.
	\label{eq-gradient}
\end{equation}

The total loss for learning synthetic images is
\begin{equation}
	\mathcal{L}_{syn} = \mathcal{L}_g + \mathcal{L}_c
	\label{eq-loss}
\end{equation}

Finally, we illustrate the overall training process of our proposed UDD in \cref{alg:pipeline}.

\section{Experiments}
\label{sec:Experiments}
In this section, we use 5 digit recognition or image classification datasets including MNIST \cite{mnist}, FashionMNIST \cite{fashion}, SVHN \cite{svhn}, CIFAR-10 and CIFAR-100 \cite{cifar} to evaluate our method. We compare the proposed method against several dataset distillation methods including Dataset Distillation (DD) \cite{DD}, Dataset Condensation (DC) \cite{DC}, Aligning Features (CAFE) \cite{CAFE}, Matching Training Trajectories (MTT) \cite{MTT} and Efficient Synthetic-Data Parameterization (IDC) \cite{IDC}, along with several coreset selection methods including random selection (Random), herding methods (Herding) \cite{subset-super-sample}, and example forgetting (Forgetting) \cite{example-forget}. We also provide the visualizations of synthetic images of our method in \cref{fig-svhn-cf10} to show the efficient utilization of the synthetic images. We report the main result by using data jittering-based policy. Due to page limited, more visualizations, ablation experiments and observation are demonstrated in the supplementary material. 

\begin{table*}[!htbp]
	\caption{Top-1 accuracy of test networks trained on synthetic images from 5 datasets on ConvNet-3. 
	As usual, we randomly initialize the network at each iteration, continuously updating a specific amount of synthetic data for each category.
	DD\textsuperscript{$\dagger$} denotes that the earlier work DD used LeNet \cite{LeNet} for MNIST and AlexNet \cite{AlexNet} for CIFAR-10. IPC means the number of synthetic images per class and Ratio means the ratio of the synthetic images to the whole original real dataset.}
	\label{tb-main}
	\scriptsize
	\centering
 \resizebox{0.99\linewidth}{!}{
	\begin{tabular}{ccc|ccc|cccccc|c}
		\toprule
		\multirow{2}*{Dataset} & \multirow{2}*{IPC} & \multirow{2}*{Ratio \%} & \multicolumn{3}{c|}{Coreset Selection} & \multicolumn{6}{c|}{Condensation Methods} & \multirow{2}*{Full Dataset} \\ 
		& & & Random & Herding & Forgetting & DD\textsuperscript{$\dagger$}  & DC  & CAFE  & MTT & IDC  & Ours \\
		\midrule
		\multirow{3}*{MNIST} 
		& 1  & 0.017 & 64.9$\pm$3.5 & 89.2$\pm$1.6 & 35.5$\pm$5.6 & -            & 91.7$\pm$0.5 & 93.1$\pm$0.3 & - & 94.2 & \textbf{95.6$\pm$0.3} & \multirow{3}*{99.6$\pm$0.0} \\
		& 10 & 0.17  & 95.1$\pm$0.9 & 93.7$\pm$0.3 & 68.1$\pm$3.3 & 79.5$\pm$8.1 & 97.4$\pm$0.2 & 97.2$\pm$0.2 & - & 98.4 & \textbf{98.4$\pm$0.3} \\
		& 50 & 0.83  & 97.9$\pm$0.2 & 94.8$\pm$0.2 & 88.2$\pm$1.2 & -            & 98.8$\pm$0.2 & 98.6$\pm$0.2 & - & 99.1 & \textbf{99.1$\pm$0.2} \\
		\midrule
		\multirow{3}*{FashionMNIST} 
		& 1  & 0.017 & 51.4$\pm$3.8 & 67.0$\pm$1.9 & 42.0$\pm$5.5 & - & 70.5$\pm$0.6 & 77.1$\pm$0.9 & - & 81.0 & \textbf{82.3$\pm$0.6} & \multirow{3}*{93.5$\pm$0.1} \\
		& 10 & 0.17  & 73.8$\pm$0.7 & 71.1$\pm$0.7 & 53.9$\pm$2.0 & - & 82.3$\pm$0.4 & 83.0$\pm$0.4 & - & 86.0 & \textbf{86.6$\pm$0.4} \\
		& 50 & 0.83  & 82.5$\pm$0.7 & 71.9$\pm$0.8 & 55.0$\pm$1.1 & - & 83.6$\pm$0.4 & 84.8$\pm$0.4 & - & 86.2 & \textbf{86.8$\pm$0.3} \\
		\midrule
		\multirow{3}*{SVHN} 
		& 1  & 0.014 & 14.6$\pm$1.6 & 20.9$\pm$1.3 & 12.1$\pm$1.7 & - & 31.2$\pm$1.4 & 42.6$\pm$3.3 & - & 68.5 & \textbf{70.1$\pm$0.4} & \multirow{3}*{95.4$\pm$0.1} \\
		& 10 & 0.14  & 35.1$\pm$4.1 & 50.5$\pm$3.3 & 16.8$\pm$1.2 & - & 76.1$\pm$0.6 & 75.9$\pm$0.6 & - & 87.5 & \textbf{89.3$\pm$0.3} \\
		& 50 & 0.7   & 70.9$\pm$0.9 & 72.6$\pm$0.8 & 27.2$\pm$1.5 & - & 82.3$\pm$0.3 & 81.3$\pm$0.3 & - & 90.1 & \textbf{90.7$\pm$0.3} \\
		\midrule
		\multirow{3}*{CIFAR-10} 
		& 1  & 0.02 & 14.4$\pm$2.0 & 21.5$\pm$1.2 & 13.5$\pm$1.2 & -            & 28.3$\pm$0.5 & 30.3$\pm$1.1 & 46.3$\pm$0.8 & 50.6 & \textbf{54.6$\pm$0.4} & \multirow{3}*{88.1$\pm$0.1} \\
		& 10 & 0.2  & 26.0$\pm$1.2 & 31.6$\pm$0.7 & 23.3$\pm$1.0 & 36.8$\pm$1.2 & 52.1$\pm$0.5 & 46.3$\pm$0.6 & 65.3$\pm$0.7 & 67.5 & \textbf{68.9$\pm$0.5} \\
		& 50 & 1    & 43.4$\pm$1.0 & 40.4$\pm$0.6 & 23.3$\pm$1.1 & -            & 60.6$\pm$0.5 & 55.5$\pm$0.6 & 71.6$\pm$0.2 & 74.5 & \textbf{75.4$\pm$0.1} \\
		\midrule
		\multirow{3}*{CIFAR-100} 
		& 1  & 0.2 & 4.2$\pm$0.3  & 8.4$\pm$0.3  & 4.5$\pm$0.3 & - & 12.8$\pm$0.3 & 12.9$\pm$0.3 & 24.3$\pm$0.3 & - & \textbf{27.9$\pm$0.4} & \multirow{3}*{56.2$\pm$0.3} \\
		& 10 & 2   & 14.6$\pm$0.5 & 17.3$\pm$0.3 & 9.8$\pm$0.2 & - & 25.2$\pm$0.3 & 27.8$\pm$0.3 & 40.1$\pm$0.4 & 44.8 & \textbf{45.8$\pm$0.3} \\
		& 50 & 10  & 30.0$\pm$0.4 & 33.7$\pm$0.5 & -           & - & -            & 37.9$\pm$0.3 & 47.7$\pm$0.2 & - & \textbf{51.3$\pm$0.3} \\
		\bottomrule
	\end{tabular}}
\vspace{-0.5cm}
\end{table*}

\subsection{Compare to State-of-the-art Methods}
As shown in \cref{tb-main}, our method achieves state-of-the-art results with all the IPC settings(IPC=1/10/50) on all 5 datasets. 
Specifically, when a smaller image per class comes to us, e.g., with IPC=1, the improvements on MNIST, FashionMNIST, SVHN, CIFAR-10 and CIFAR-100 are 1.4\%, 1.3\%, 1.6\%, 4.0\% and 3.7\% over the next best method (IDC \cite{IDC} and MTT \cite{MTT}). Compared with the coreset method, our method outperforms Random, Herding, and Forgetting by a large margin.
When using 10 and 50 synthetic images per class, the advantage of our method drops slightly but still outperforms other methods. 
In general, our method performs better in more demanding scenarios (smaller IPC or harder datasets like CIFAR-10/100).
This is because demanding scenes require condensing more useful information in fewer images, thus the synthetic images should be more informative and discriminate, namely higher utilization, which is the focus of our method.

\subsection{Ablation Studies}
\label{sec:ablation}
In this section, we perform ablation studies on our method using the CIFAR-10 dataset with IPC = 10.

\noindent\textbf{Evaluation of Underutilized Regions Search Policy.} 
\label{ab_policy}
We conduct ablation experiments to explore the effect of each underutilized region searching approach. 
\cref{tb-three-ud} shows the consistent performance gain with those three underutilized region searching approaches. 
When using one of these methods alone, data jittering has a better performance but is relatively time-consuming. 
When employing the three approaches together to determine the underutilized regions by voting, the performance gains the most while significant time cost increases. Therefore, we use data jittering alone as the underutilized region searching approach in our method.

\begin{table}[!htbp]
	\caption{Ablation experiments on applying three underutilized region searching approaches. Time: the iter time it takes to perform searching on one synthetic image.}
	\label{tb-three-ud}
	\small
	\centering
	\begin{tabular}{ccc|c|c}
		\toprule
		data jittering & gradient & activation & Accuracy (\%) & Time (ms)\\
		\midrule
		\checkmark &&& 68.9 & 17.2\\
		&\checkmark&& 68.1 & 14.6\\
		&&\checkmark& 68.3 & 15.8\\
		\checkmark&\checkmark&\checkmark& 69.1 & 20.1\\
		\bottomrule
	\end{tabular}
	\vspace{-0.5cm}
\end{table}

\noindent\textbf{Evaluation of Jitter Function.}
In \cref{tb-jitter}, we explore the effect of the jitter function used in underutilized region searching.
We find that using different jitter functions yielded little difference in the final performances, which means that the choice of jitter function is not a key factor affecting our method.

\begin{table}[!htbp]
	\caption{Ablation experiment on jitter functions.}
	\label{tb-jitter}
	\small
	\centering
	\begin{tabular}{c|c|c|c}
		\toprule
		Jitter Function & uniform & salt and pepper & gaussian \\
		\midrule
		Accuracy (\%) & 68.6 & 68.7 & 68.9 \\
		\bottomrule
	\end{tabular}
\vspace{-0.5cm}
\end{table}

\noindent\textbf{Evaluation of CFC loss.} 
To measure the effectiveness of the CFC loss, we conduct the ablation experiments in \cref{tb-cfc}. By applying CFC loss, both IDC and our method have some performance gains.

\begin{table}[!htbp]
	\caption{Ablation experiment on applying the category-wise feature contrastive (CFC) loss.}
	\label{tb-cfc}
	\small
	\centering
	\begin{tabular}{c|cc|cc}
		\toprule
		& \multicolumn{2}{c|}{IDC} & \multicolumn{2}{c}{Ours} \\
		\midrule
		+CFC & & \checkmark & & \checkmark \\
		\midrule
		Accuracy (\%) & 67.5 & 67.8 & 68.5 & 68.9 \\
		\bottomrule
	\end{tabular}
\vspace{-0.7cm}
\end{table}
\begin{figure}[h]
	\centering
	\includegraphics[width=0.6\linewidth]{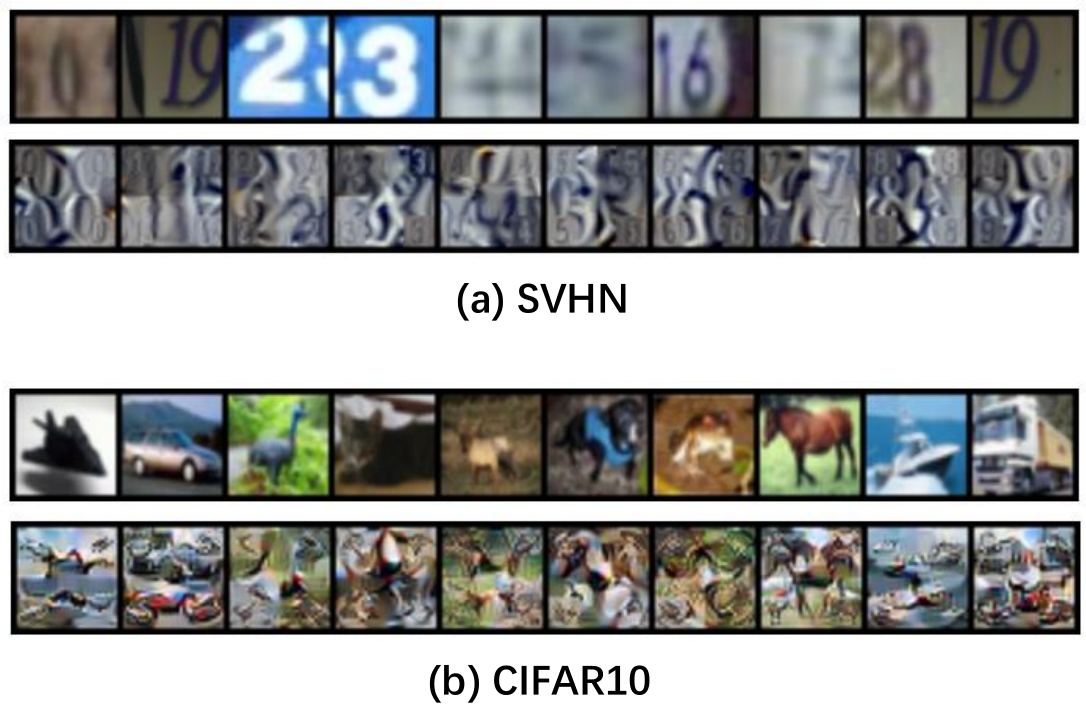}
	\caption{The synthetic datasets generated by our method on SVHN and CIFAR-10 with IPC=1.
		The top row of each sub-image is the real samples.}
	\label{fig-svhn-cf10}
\vspace{-0.5cm} 
\end{figure}

\subsection{Visualizations}
We provide visualizations of the synthetic dataset on SVHN and CIFAR-10 in \cref{fig-svhn-cf10}. 
Compared to the real images, our synthetic images have a significant advantage in the number and size diversity of foreground objects.

\section{Conclusion}
In this paper,  we propose a novel approach to dataset distillation that focuses on optimizing the underutilized regions to enhance the semantic information of synthetic images. Our method includes two underutilized regions searching policies, the response-based policy and the data jittering-based policy, that are designed for different conditions. Additionally, we apply a category-wise feature contrastive loss to increase the distinguishability of different categories in the synthetic dataset. Our extensive experimental results on various datasets demonstrate that our method consistently outperforms state-of-the-art methods. In future work, we plan to investigate more optimal ways to distribute the number of synthetic images across different categories.

\vspace{1em}
\noindent\textbf{Acknowledgement.}
This work was supported  by the NNSFC\&CAAC under Grants U2233209, and in part by the National Natural Science Foundation of China under Grant 62071104, in part by the Natural Science Foundation of Sichuan, China under Grant 2023NSFSC0484, in part by the NNSFC\&CAAC under Grants U2133211.
%
%
%
\bibliographystyle{splncs04}
\bibliography{egbib}
%




\clearpage
\input{appendix}
\end{document}

%% file: appendix.tex
%
%
\renewcommand\UrlFont{\color{blue}\rmfamily}


%
\appendix

\section{Datasets and Implementation Details}
\noindent\textbf{MNIST} \cite{mnist}. The MNIST dataset is a large-scale handwritten digit database, containing a training set of 60,000 examples and a test set of 10,000 examples with a size of 28 $\times$ 28.

\noindent\textbf{FashionMNIST} \cite{fashion}. FashionMNIST covers 70,000 different product front images from 10 categories with the size of 28 $\times$ 28, which are divided into a training set of 60,000 and a test set of 10,000. 

\noindent\textbf{SVHN} \cite{svhn}. SVHN dataset is a collection of house numbers from Google street view, which contains over 600,000 labeled digit images with the size of 32 $\times$ 32.

\noindent\textbf{CIFAR-10/100} \cite{cifar}. CIFAR-10 and CIFAR-100 are labeled subsets of the 80 million tiny image dataset. The two CIFAR datasets contain a training set of 50,000 and a test set of 10,000 with the size of 32 $\times$ 32.

\noindent\textbf{Implementation Details.}
We follow the common settings proposed by DD \cite{DD} to evaluate our method. Specifically, the performance of the synthetic data is measured by the test accuracy of the neural networks trained on the synthetic data. We optimize 1/10/50 Images Per Class (IPC) for each experiment of the above 5 datasets using a three-layer Convolutional Network (ConvNet-3), which includes three repeated convolutional blocks.
The hyper-parameters are set as $N = 12$, $M = 2$, $P = 8$, $\tau = 0.7$ and $\alpha = 0.5$.
The initial learning rate for the synthetic image is 0.005, which is divided by 2 in 1200, 1600, and 1800 iterations with the whole training iterations of 2000.
The neural networks are trained on real images with a fixed learning rate of 0.01 for 300 epochs.

\section{More Ablation Studies}
\label{sec:ablation1}
In this section, we perform more ablation studies on our method using the CIFAR-10 dataset with IPC = 10.

\noindent\textbf{Evaluation of Candidate Regions.} 
$N$ is the number of candidate regions of each synthetic data.
We study different values of $N$, which denotes the combination of candidate regions of different sizes. 
Specifically, 
given a synthetic image with size of $\mathit{H} \times \mathit{W}$, candidate regions have the dimensions of $(\mathit{H}/3\times\mathit{W}/3),(\mathit{H}/2\times\mathit{W}/2)$ and $(2\mathit{H}/3\times2\mathit{W}/3)$ separately. $N$ = 4, 8, and 12 represent choose one, two, or three types of regions respectively. We fix the ratio of $P$ to $M$ in implementation.
As shown in \cref{tb-hyper-N}, the best performance can be achieved by using all three types of regions at the same time,
which is friendly to the condensation of small, medium, and large objects and supports information sharing in various regions.

\begin{table}[!htbp]
	\caption{Evaluation of Candidate Regions.}
	\label{tb-hyper-N}
	\small
	\centering
	\begin{tabular}{c|ccc|c}
		\toprule
		$N$ & 1/3 & 1/2 & 2/3 &  Accuracy (\%)\\
		\midrule
		\multirow{3}*{4} &\checkmark &&& 66.4  \\
		&&\checkmark&& 67.5 \\
		&	&&\checkmark& 67.8 \\
		\midrule
		\multirow{3}*{8} &\checkmark &\checkmark&& 67.9 \\
		&\checkmark&&\checkmark& 68.0 \\
		&&\checkmark&\checkmark& 68.3 \\
		\midrule
		12 &\checkmark&\checkmark&\checkmark&  68.9 \\
		\bottomrule
	\end{tabular}
\vspace{-0.5cm}
\end{table}
\noindent\textbf{Cross-Architecture Generalization.} Synthetic dataset trained on the architecture that is used in the synthesis process usually performs better, while other architectures suffer from much over-fitting problems \cite{MTT}. To evaluate the generalization ability of the synthetic images condensed by our method, we evaluate the performance of synthetic images trained with ConvNet on other models (AlexNet \cite{AlexNet}, VGG \cite{VGG} and ResNet \cite{ResNet}). As shown in \cref{tb-general}, our method is less sensitive to architectural changes and achieves better generalization performance than other methods. 
\begin{table}[!htbp]
	\caption{Top-1 accuracy (\%) on different architectures. The synthetic images are learned on ConvNet with IPC = 10.}
	\label{tb-general}
	\small
	\centering
	\begin{tabular}{c|cccc}
		\toprule
		\multirow{2}*{} & \multicolumn{4}{c}{Evaluation Model} \\
		& ConvNet & RseNet18 & VGG11 & AlexNet \\
		\midrule
		KIP  & 47.6 & 36.8 & 42.1 & 24.4 \\
		MTT  & 64.3 & 46.4 & 50.3 & 34.2 \\
		IDC  & 67.5 & 62.7 & 63.5 & 64.2 \\
		Ours & \textbf{68.9} & \textbf{64.1} & \textbf{63.8} & \textbf{66.3} \\
		\bottomrule
	\end{tabular}
	\vspace{-0.4cm}
\end{table}
\section{Compared with KIP}
Compared with Kernel Inducing Point (KIP) \cite{Infinitely} that uses neural tangent kernel for training networks, we follow the test settings of \cite{MTT} to test the results on both a ConvNet with the width of 1024 (the original setting in KIP paper) and a ConvNet with the width of 128 (same as our and other condensation methods).
As shown in \cref{tb-kip}, our method outperforms KIP in all settings with a large margin.

\begin{table}[h]
	\caption{Top-1 accuracy of networks trained on the synthetic images using KIP and our method.}
	\label{tb-kip}
	\scriptsize
	\centering
	\begin{tabular}{ccc|c|cc}
		\toprule
		\multirow{2}{*}{} & \multirow{2}{*}{IPC}  & \multirow{2}{*}{Ratio \%} & KIP to NN & KIP to NN & Ours \\
		& & & (1024-width) & (128-width) & (128-width) \\
		\midrule
		\multirow{3}*{CIFAR-10}
		& 1  & 0.02 & 49.9 & 38.3 & \textbf{54.6} \\
		& 10 & 0.2  & 62.7 & 57.6 & \textbf{68.9}\\
		& 50 & 1    & 68.6 & 65.8 & \textbf{75.4} \\
		\midrule
		\multirow{2}*{CIFAR-100}
		& 1  & 0.2 & 15.7 & 18.2 & \textbf{27.9} \\
		& 10 & 2   & 28.3 & 32.8 & \textbf{45.8} \\
		\bottomrule
	\end{tabular}
\vspace{-0.3cm}
\end{table}

\section{Experiments on Higher Resolution}
We also implement our method on higher resolution images in the form of 224 $\times$ 224 subsets of ImageNet \cite{imagenet}.
We follow the evaluation settings proposed by IDC \cite{IDC}, which use 
a ten-layer ResNet \cite{ResNet} with Averagepooling (ResNetAp-10) as the training network
and a subclass list from CMC \cite{CMC} that containing 10 classes
with 10/20 Images Per Class (IPC).
\cref{tb-imagenet} shows that our method outperforms the existing methods on the higher resolution dataset.

\begin{table}[!htbp]
	\caption{Top-1 accuracy of networks trained on synthetic images condensed from ImageNet-subset. Accuracy of the original subset with class = 10 is 90.8\%.}
	\label{tb-imagenet}
	\centering
	\begin{tabular}{ccc|cccc}
		\toprule
		Class & IPC & Ratio (\%) & DSA & DM  & IDC  & Ours \\
		\midrule
		\multirow{2}{*}{10} & 10 & 0.8 & 52.7 & 52.3 & 72.8 & \textbf{73.2} \\
		& 20 & 1.6 & 57.4 & 59.3 & 76.6 & \textbf{77.1} \\
		\bottomrule
	\end{tabular}
\vspace{-0.3cm}
\end{table}

\section{Observation}
\begin{figure}[!htbp]
	\centering
	\includegraphics[width=0.55\linewidth]{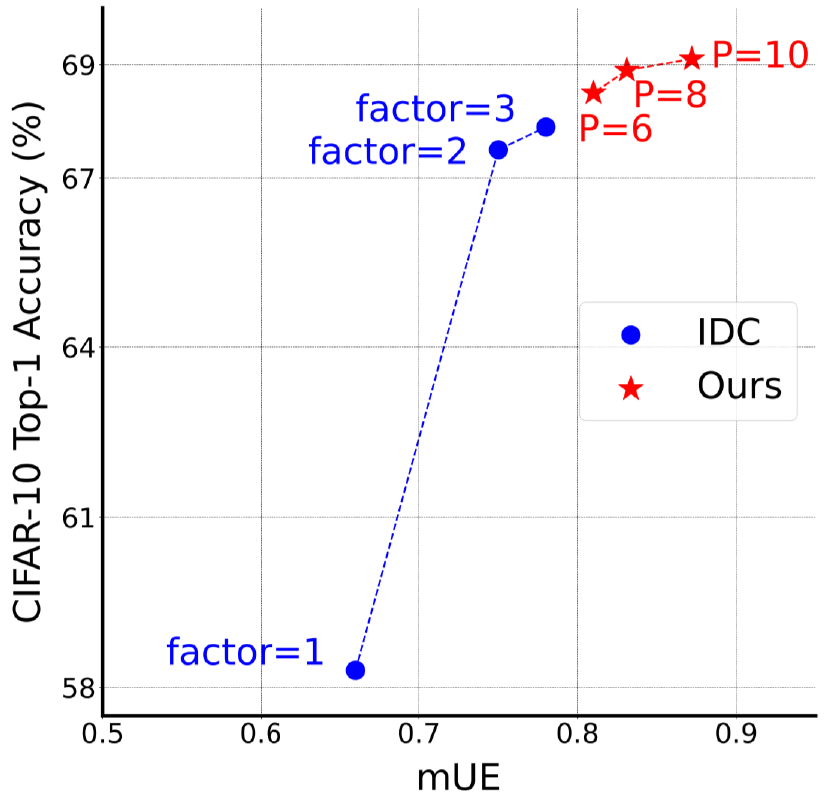}
	\caption{Accuracy of ConvNet-3 trained on the synthetic dataset with increasing utilization.
     $D$ start from 5 to 10 with a step size of 1.
	}
	\label{fig-utilization}
\end{figure}

\begin{figure}[!htbp]
	\centering
\includegraphics[width=0.55\linewidth]{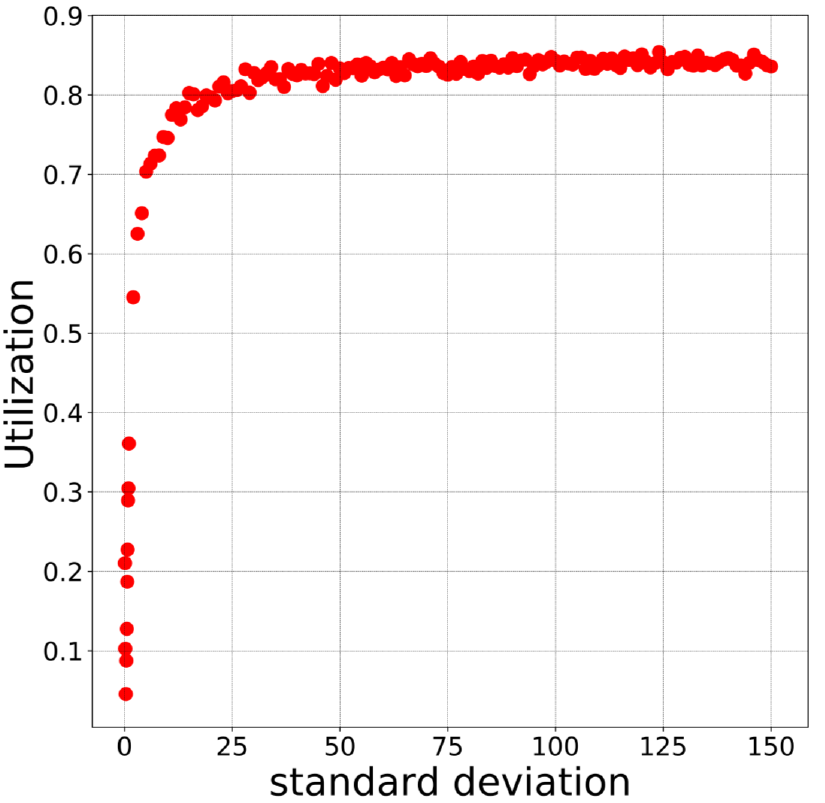}
	\caption{Relationship between distribution and utilization.}
	\label{fig-std-ue}
\vspace{-0.3cm}
\end{figure}

\label{obser}
In this section, we try to quantify the utilization ratio of synthetic data by analyzing the distribution of activations. We also provide empirical observations on the relationship between accuracy and the utilization of synthetic data.

\begin{figure}[!htbp]
	\centering
\includegraphics[width=0.95\linewidth]{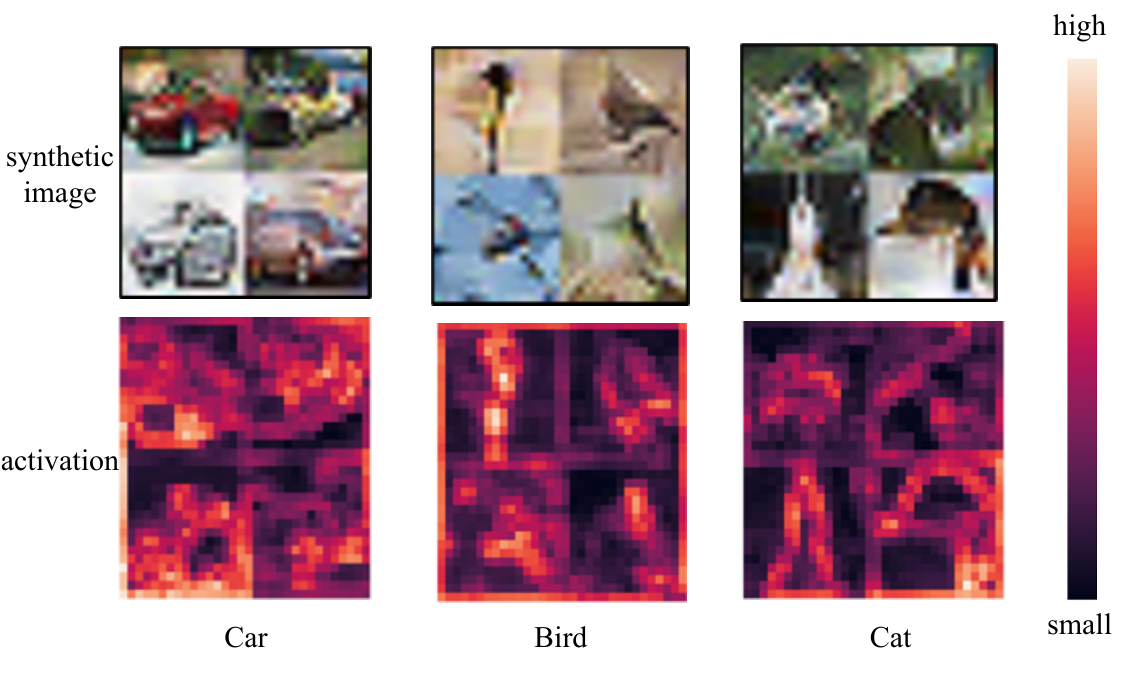}
		\caption{Examples of the synthetic images with their activation maps from 32 $\times$ 32 CIFAR-10 dataset. The synthetic images are generated by IDC \cite{IDC}.}
		\label{fig:fig-act1}
\end{figure}
As shown in \cref{fig:fig-act1}, activations can indicate the level of response of the network to each region of the synthetic images. A more concentrated activation distribution indicates that more regions are underutilized.
We hypothesize that a synthetic dataset with better utilization, which has fewer underutilized regions, will produce a more balanced activation distribution, i.e., a smaller entropy between activations.
Meanwhile, the synthetic dataset with higher utilization can condense more valuable information, leading to better performance.
Thus, we design a new evaluation metric, called $mUE$ (mean Utilization measured by activation Entropy), which is computed as follows:
for a given synthetic image, we divide its activation map $X$ into $d$ non-overlapping but equal-sized intervals, and compute the distribution $p(X^d)$ of activations in each interval. We repeat this process for multiple partitions $D$ and compute the mean of the entropy over all partitions:
\begin{equation}
	UE^{d} = 1 + \frac{1}{|X|log(d)} \sum_{i={\frac{1}{d}}}^{d} p(X^d)_i log(p(X^d)_i), 
	\label{eq-u-entropy}
\end{equation}
\begin{equation}
	mUE = \frac{1}{|D|} \sum_{d \in D}UE^{d},
	\label{eq-mue}
\end{equation}
where $p(X^d)_i$ is the probability of the $i$-th interval, computed as the ratio of the number of activations falling in that interval to the total number of activations in the image. A higher $mUE$ value indicates a more balanced distribution of activations across intervals and therefore better utilization of the synthetic image.

\cref{fig-utilization} shows the performance of synthetic dataset with different $mUE$ of 10 images per class on CIFAR-10.
We change hyperparameters (\emph{factor} in IDC and $P$ in our method) to obtain multiple results of IDC and our method.
The figure shows a positive correlation between $mUE$ and accuracy, and our method outperforms IDC with higher accuracy and $mUE$ under all settings.

\noindent\textbf{Relationship between Distribution and Utilization.}
We use the entropy of the distribution of activations to estimate the utilization of the synthetic dataset.
To show the relationship between distribution and $UE$ more intuitively, we generate various data subjected to a normal distribution and calculate their utilization measured by entropy. Specifically, we obtain different distributions by adjusting the standard deviation (std).

As shown in \cref{fig-std-ue}, when we increase std of the normal distribution, utilization becomes larger firstly and then grows slowly. The largest utilization value brought by increasing std of normal distribution is 0.86, which is numerically close to our method.

\section{More Visualizations}
\begin{figure}[!htbp]
	\centering
	\includegraphics[width=0.85\linewidth]{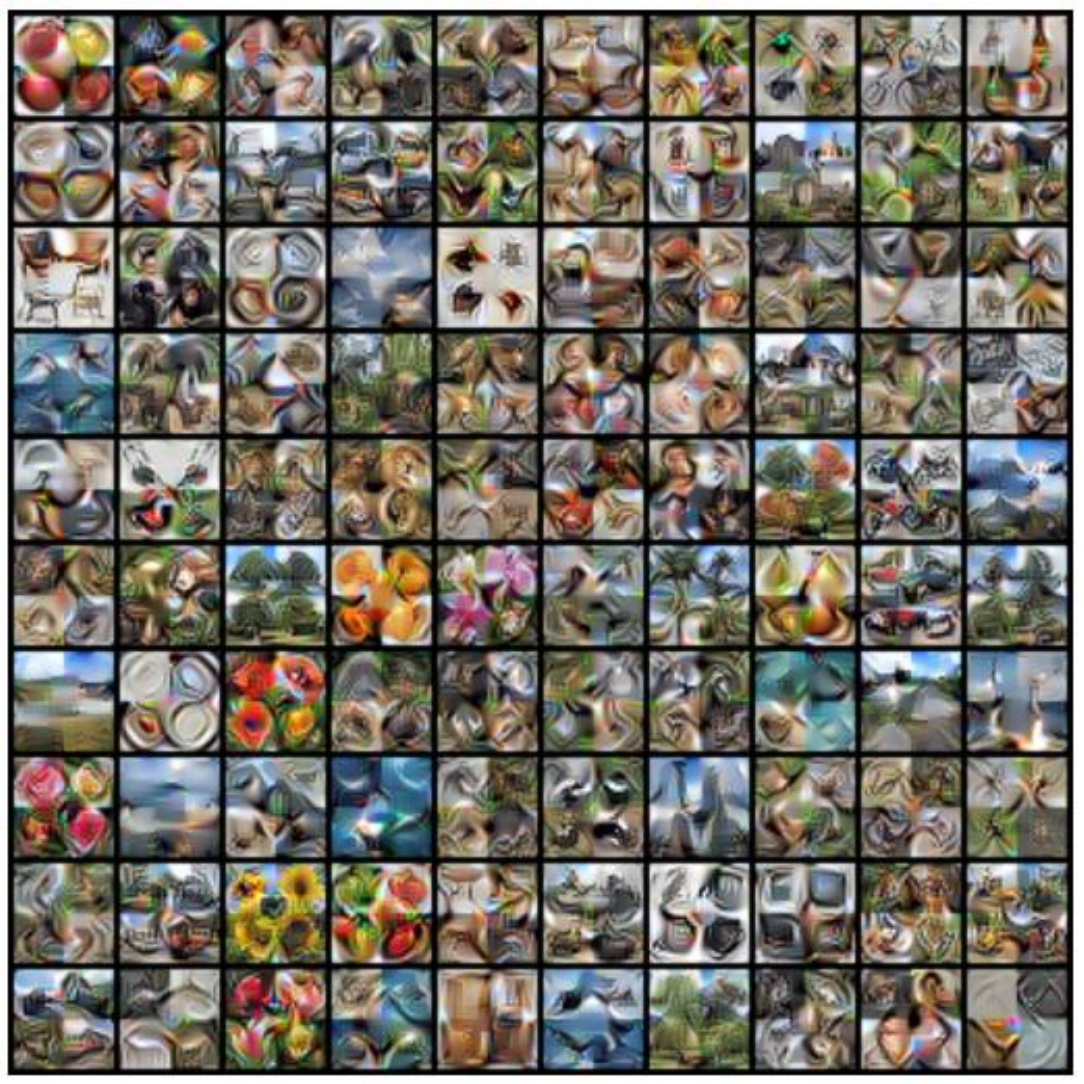}
	\caption{The synthetic dataset generated by our method on CIFAR-100 with IPC=1.}
	\label{fig-cf100}
\vspace{-0.3cm}
\end{figure}

\begin{figure*}[!htbp]
	\centering
	\includegraphics[width=0.95\linewidth]{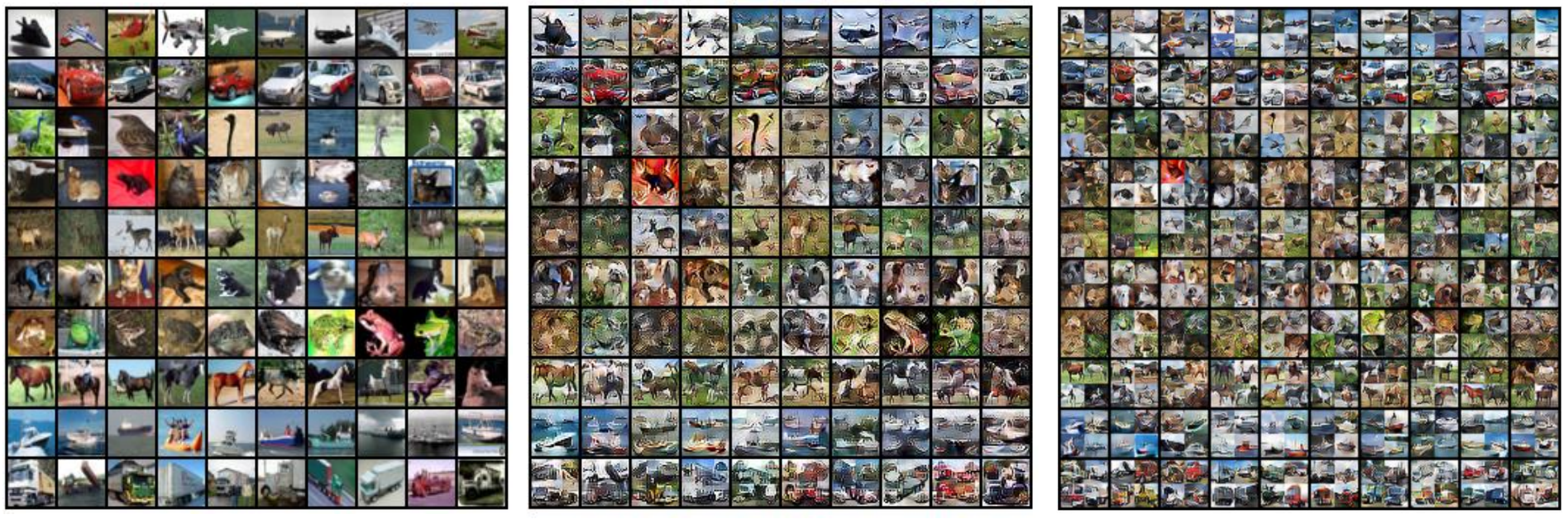}
	\caption{Visualizations of images from CIFAR-10 with IPC = 10.
		\textbf{Left}: 10 random selections from each category of original images.
		\textbf{Middle}: The synthetic images of our method.
		\textbf{Right}: The synthetic images of IDC.
	}
	\label{fig-visual-cifar}
	\vspace{-0.3cm}
\end{figure*}

We also provide visualizations of the synthetic dataset on CIFAR-10 and CIFAR-100 in \cref{fig-visual-cifar} and \cref{fig-cf100} under different IPC settings.

\section{Related Work}
\noindent\textbf{Dataset Distillation.} 
Wang et al. \cite{DD} first introduce dataset distillation in the image classification task to condense information from a large dataset into a smaller synthetic dataset.
Subsequently, continued improvements to the optimization scheme in the synthesis process lead to better results. 
\cite{Soft-Label,Flexible} increase the available information for model training by synthesizing data and distilling labels simultaneously, obtaining a significant performance boost. Furthermore, several methods are verified to be effective, including gradient matching \cite{DC}, data augmentations \cite{DSA}, feature alignment \cite{CAFE,liu2023dream}, training trajectories matching \cite{MTT}, subset matching \cite{du2024sequential}, and optimizing for the infinite-width kernel limit \cite{Infinitely,Kernel}. 
Shin et al. \cite{shin2024frequency} proposed a frequency domain-based dataset distillation approach, exploring new dimensions for data synthesis. Li et al.\cite{li2024importance} presented an importance-aware adaptive dataset distillation technique, focusing on optimizing the importance of distilled data. Moser et al. \cite{moser2024latent} introduced latent dataset distillation using diffusion models, a new frontier in the distillation process that utilizes the potential of latent space representations.
Dataset distillation has achieved impressive results in various applications including cross-dataset task \cite{Flexible}, text distillation \cite{Soft-Label}, continual learning \cite{DD,DC,DSA}, neural architecture search \cite{DC,DSA}, and federated learning \cite{SecDD,FedSynth,One-Shot}. For a comprehensive surveys and new methodologies, we refer the readers to \cite{lei2023comprehensive,yu2023dataset}.
IDC \cite{IDC} first improves the synthetic set performance by efficiently utilizing regions of synthetic images. 
However, it still suffers from fixed and independent partitioning of each synthetic image into sub-images. 
Thus handcraft design can not able to make full use of each region of the synthetic image, while our utilization-sensitive process overcomes this limitation by dynamically searching these underutilized regions and improving them via gradient optimization.

\noindent\textbf{Knowledge Distillation.} 
The original proposal of dataset distillation is inspired by knowledge distillation \cite{KD}, which is one of the representative techniques of model compression \cite{Model-compression}. 
Knowledge distillation methods have been extensively studied in recent years, forming three basic knowledge categories: response-based knowledge \cite{KD,LEOD,Conditional}, feature-based knowledge \cite{Fitnets,ABLoss,OFD} and relation-based knowledge \cite{Gift,RKD}. 
Response-based knowledge usually refers to the neural response of the last output layer of the teacher model, which is known as soft-label.
Similarly, feature-based knowledge distillation and relation-based knowledge distillation continuously mine the effective information of the middle-level layer of the teacher model.
The above knowledge distillation methods are dedicated to constructing more effective knowledge from the fixed teacher model to improve the performance of the student model.
Our method also emphasizes better utilization of limited resources, i.e., the limited area of the synthetic images.

\noindent\textbf{Subset Selection.}
The classic approach to implement dataset distillation is subset selection \cite{subset-super-sample,subset-approximating,coreset,online-coreset,coreset-kmeans,coreset-logistic-regression,coreset-scalable}, which aims to construct a valuable subset from the entire training dataset, where training on this small subset achieves good performance.
Most of the existing constructions were limited to simple models such as k-means \cite{coreset-kmeans} and logistic regression \cite{coreset-logistic-regression}. Several approaches relax the uniform approximation \cite{coreset-hilbert,coreset-bi-level} or aim to filter samples that are more valuable to the training process, including measuring single-example accuracy \cite{subset-valuable} and counting misclassification rates \cite{example-forget}.
However, these methods require a large number of training samples to construct a better subset, in part because their valuable images have to be real, whereas the condensed images do not need to be realistic and exempt from this constraint.

%





%% file: main.bbl
\begin{thebibliography}{10}
\providecommand{\url}[1]{\texttt{#1}}
\providecommand{\urlprefix}{URL }
\providecommand{\doi}[1]{https://doi.org/#1}

\bibitem{subset-approximating}
Agarwal, P.K., Har{-}Peled, S., Varadarajan, K.R.: Approximating extent measures of points. ACM pp. 606--635 (2004)

\bibitem{Flexible}
Bohdal, O., Yang, Y., Hospedales, T.M.: Flexible dataset distillation: Learn labels instead of images. arXiv preprint arXiv:2006.08572  (2020)

\bibitem{borji2019salient}
Borji, A., Cheng, M.M., Hou, Q., Jiang, H., Li, J.: Salient object detection: A survey. Computational visual media  \textbf{5},  117--150 (2019)

\bibitem{borji2015salient}
Borji, A., Cheng, M.M., Jiang, H., Li, J.: Salient object detection: A benchmark. IEEE transactions on image processing  \textbf{24}(12),  5706--5722 (2015)

\bibitem{coreset-bi-level}
Borsos, Z., Mutny, M., Krause, A.: Coresets via bilevel optimization for continual learning and streaming. In: NIPS (2020)

\bibitem{Model-compression}
Bucila, C., Caruana, R., Niculescu{-}Mizil, A.: Model compression. In: KDD. pp. 535--541 (2006)

\bibitem{coreset-hilbert}
Campbell, T., Broderick, T.: Automated scalable bayesian inference via hilbert coresets. J. Mach. Learn. Res. pp. 15:1--15:38 (2019)

\bibitem{MTT}
Cazenavette, G., Wang, T., Torralba, A., Efros, A.A., Zhu, J.: Dataset distillation by matching training trajectories. In: CVPR. pp. 10708--10717 (2022)

\bibitem{LEOD}
Chen, G., Choi, W., Yu, X., Han, T.X., Chandraker, M.: Learning efficient object detection models with knowledge distillation. In: NIPS. pp. 742--751 (2017)

\bibitem{SimCLR}
Chen, T., Kornblith, S., Norouzi, M., Hinton, G.E.: A simple framework for contrastive learning of visual representations. In: ICML. pp. 1597--1607 (2020)

\bibitem{subset-super-sample}
Chen, Y., Welling, M., Smola, A.: Super-samples from kernel herding. In: Proceedings of the Twenty-Sixth Conference on Uncertainty in Artificial Intelligence. pp. 109--116 (2010)

\bibitem{imagenet}
Deng, J., Dong, W., Socher, R., Li, L.J., Li, K., Fei-Fei, L.: Imagenet: A large-scale hierarchical image database. In: CVPR. pp. 248--255. Ieee (2009)

\bibitem{mnist}
Deng, L.: The mnist database of handwritten digit images for machine learning research. IEEE Signal Processing Magazine p. 29(6):141–142 (2012)

\bibitem{du2024sequential}
Du, J., Shi, Q., Zhou, J.T.: Sequential subset matching for dataset distillation. Advances in Neural Information Processing Systems  \textbf{36} (2024)

\bibitem{coreset-scalable}
Feldman, D., Faulkner, M., Krause, A.: Scalable training of mixture models via coresets. In: NIPS. pp. 2142--2150 (2011)

\bibitem{coreset-kmeans}
Feldman, D., Langberg, M.: A unified framework for approximating and clustering data. In: ACM. pp. 569--578 (2011)

\bibitem{FLSD}
Goetz, J., Tewari, A.: Federated learning via synthetic data. arXiv preprint arXiv:2008.04489  (2020)

\bibitem{Gaussian}
Gonz{\'a}lez, R.C., Woods, R.E.: Digital image processing. PAMI pp. 242--243 (1981)

\bibitem{ResNet}
He, K., Zhang, X., Ren, S., Sun, J.: Deep residual learning for image recognition. In: CVPR. pp. 770--778 (2016)

\bibitem{OFD}
Heo, B., Kim, J., Yun, S., Park, H., Kwak, N., Choi, J.Y.: A comprehensive overhaul of feature distillation. In: ICCV. pp. 1921--1930 (2019)

\bibitem{ABLoss}
Heo, B., Lee, M., Yun, S., Choi, J.Y.: Knowledge transfer via distillation of activation boundaries formed by hidden neurons. In: AAAI. pp. 3779--3787 (2019)

\bibitem{KD}
Hinton, G., Vinyals, O., Dean, J.: Distilling the knowledge in a neural network. stat  \textbf{1050}, ~9 (2015)

\bibitem{FedSynth}
Hu, S., Goetz, J., Malik, K., Zhan, H., Liu, Z., Liu, Y.: Fedsynth: Gradient compression via synthetic data in federated learning. arXiv preprint arXiv:2204.01273  (2022)

\bibitem{coreset-logistic-regression}
Huggins, J.H., Campbell, T., Broderick, T.: Coresets for scalable bayesian logistic regression. In: NIPS. pp. 4080--4088 (2016)

\bibitem{IDC}
Kim, J., Kim, J., Oh, S.J., Yun, S., Song, H., Jeong, J., Ha, J., Song, H.O.: Dataset condensation via efficient synthetic-data parameterization. In: ICML. pp. 11102--11118 (2022)

\bibitem{cifar}
Krizhevsky, A., Hinton, G., et~al: Learning multiple layers of features from tiny images (2009)

\bibitem{AlexNet}
Krizhevsky, A., Sutskever, I., Hinton, G.E.: Imagenet classification with deep convolutional neural networks. In: NIPS. pp. 1106--1114 (2012)

\bibitem{subset-valuable}
Lapedriza, A., Pirsiavash, H., Bylinskii, Z., Torralba, A.: Are all training examples equally valuable. UMBC Faculty Collection  (2013)

\bibitem{LeNet}
LeCun, Y., Bottou, L., Bengio, Y., Haffner, P.: Gradient-based learning applied to document recognition. Proc. {IEEE} pp. 2278--2324 (1998)

\bibitem{lei2023comprehensive}
Lei, S., Tao, D.: A comprehensive survey of dataset distillation. IEEE Transactions on Pattern Analysis and Machine Intelligence  (2023)

\bibitem{X-Ray}
Li, G., Togo, R., Ogawa, T., Haseyama, M.: Soft-label anonymous gastric x-ray image distillation. In: ICIP. pp. 305--309 (2020)

\bibitem{li2024importance}
Li, G., Togo, R., Ogawa, T., Haseyama, M.: Importance-aware adaptive dataset distillation. Neural Networks  \textbf{172},  106154 (2024)

\bibitem{liu2023dream}
Liu, Y., Gu, J., Wang, K., Zhu, Z., Jiang, W., You, Y.: Dream: Efficient dataset distillation by representative matching. In: Proceedings of the IEEE/CVF International Conference on Computer Vision. pp. 17314--17324 (2023)

\bibitem{tsne}
Van~der Maaten, L., Hinton, G.: Visualizing data using t-sne. Journal of machine learning research  (2008)

\bibitem{GHO}
Maclaurin, D., Duvenaud, D., Adams, R.P.: Gradient-based hyperparameter optimization through reversible learning. In: ICML. pp. 2113--2122 (2015)

\bibitem{Conditional}
Meng, Z., Li, J., Zhao, Y., Gong, Y.: Conditional teacher-student learning. In: ICASSP. pp. 6445--6449 (2019)

\bibitem{moser2024latent}
Moser, B.B., Raue, F., Palacio, S., Frolov, S., Dengel, A.: Latent dataset distillation with diffusion models. arXiv preprint arXiv:2403.03881  (2024)

\bibitem{Kernel}
Nguyen, T., Chen, Z., Lee, J.: Dataset meta-learning from kernel ridge-regression. In: ICLR (2021)

\bibitem{Infinitely}
Nguyen, T., Novak, R., Xiao, L., Lee, J.: Dataset distillation with infinitely wide convolutional networks. In: NeurIPS. pp. 5186--5198 (2021)

\bibitem{RKD}
Park, W., Kim, D., Lu, Y., Cho, M.: Relational knowledge distillation. In: CVPR. pp. 3967--3976 (2019)

\bibitem{Fitnets}
Romero, A., Ballas, N., Kahou, S.E., Chassang, A., Gatta, C., Bengio, Y.: Fitnets: Hints for thin deep nets. In: ICLR (2015)

\bibitem{coreset}
Sener, O., Savarese, S.: Active learning for convolutional neural networks: {A} core-set approach. In: ICLR (2018)

\bibitem{svhn}
Sermanet, P., Chintala, S., LeCun, Y.: Convolutional neural networks applied to house numbers digit classification. In: ICPR. pp. 3288--3291 (2012)

\bibitem{shin2024frequency}
Shin, D., Shin, S., Moon, I.C.: Frequency domain-based dataset distillation. Advances in Neural Information Processing Systems  \textbf{36} (2024)

\bibitem{VGG}
Simonyan, K., Zisserman, A.: Very deep convolutional networks for large-scale image recognition. In: ICLR (2015)

\bibitem{Generative-Teaching-Networks}
Such, F.P., Rawal, A., Lehman, J., Stanley, K.O., Clune, J.: Generative teaching networks: Accelerating neural architecture search by learning to generate synthetic training data. In: ICML. pp. 9206--9216 (2020)

\bibitem{SecDD}
Sucholutsky, I., Schonlau, M.: Secdd: Efficient and secure method for remotely training neural networks (student abstract). In: AAAI. pp. 15897--15898 (2021)

\bibitem{Soft-Label}
Sucholutsky, I., Schonlau, M.: Soft-label dataset distillation and text dataset distillation. In: IJCNN. pp.~1--8 (2021)

\bibitem{CMC}
Tian, Y., Krishnan, D., Isola, P.: Contrastive multiview coding. In: ECCV. pp. 776--794 (2020)

\bibitem{example-forget}
Toneva, M., Sordoni, A., des Combes, R.T., Trischler, A., Bengio, Y., Gordon, G.J.: An empirical study of example forgetting during deep neural network learning. In: ICLR (2019)

\bibitem{CAFE}
Wang, K., Zhao, B., Peng, X., Zhu, Z., Yang, S., Wang, S., Huang, G., Bilen, H., Wang, X., You, Y.: {CAFE:} learning to condense dataset by aligning features. In: CVPR. pp. 12186--12195 (2022)

\bibitem{DD}
Wang, T., Zhu, J., Torralba, A., Efros, A.A.: Dataset distillation. arXiv preprint arXiv:1811.10959  (2018)

\bibitem{fashion}
Xiao, H., Rasul, K., Vollgraf, R.: Fashion-mnist: a novel image dataset for benchmarking machine learning algorithms. arXiv preprint arXiv:1708.07747  (2017)

\bibitem{Gift}
Yim, J., Joo, D., Bae, J., Kim, J.: A gift from knowledge distillation: Fast optimization, network minimization and transfer learning. In: CVPR. pp. 7130--7138 (2017)

\bibitem{online-coreset}
Yoon, J., Madaan, D., Yang, E., Hwang, S.J.: Online coreset selection for rehearsal-based continual learning. In: ICLR (2022)

\bibitem{yu2023dataset}
Yu, R., Liu, S., Wang, X.: Dataset distillation: A comprehensive review. IEEE Transactions on Pattern Analysis and Machine Intelligence  (2023)

\bibitem{DSA}
Zhao, B., Bilen, H.: Dataset condensation with differentiable siamese augmentation. In: ICML. pp. 12674--12685 (2021)

\bibitem{DC}
Zhao, B., Mopuri, K.R., Bilen, H.: Dataset condensation with gradient matching. In: ICLR (2021)

\bibitem{One-Shot}
Zhou, Y., Pu, G., Ma, X., Li, X., Wu, D.: Distilled one-shot federated learning. arXiv preprint arXiv:2009.07999  (2020)

\end{thebibliography}
